\pdfoutput=1
\documentclass[11pt]{article}

\usepackage{url}
\usepackage{acl}
\usepackage{booktabs}
\usepackage{CJKutf8}
\usepackage{lineno,hyperref}
\usepackage{amsfonts}
\usepackage{times}
\usepackage{latexsym}
\usepackage{amsmath}
\usepackage{amssymb}
\usepackage{enumitem}
\usepackage[T1]{fontenc}
\usepackage{graphicx} %插入图片的宏包
\usepackage{float} %设置图片浮动位置的宏包
\usepackage{subfigure} %插入多图时用子图显示的宏包
\usepackage[utf8]{inputenc}
\usepackage{multirow}
\usepackage{microtype}
\usepackage{flushend}
\usepackage{caption}
\title{Vision-Language Pre-Training for Multimodal Aspect-Based Sentiment Analysis}

\author{Yan Ling, Jianfei Yu$^{*}$, and Rui Xia\thanks{\hspace{0.3em} Corresponding authors.}\\
   School of Computer Science and Engineering, \\
   Nanjing University of Science and Technology, China \\
  \texttt{\{ylin, jfyu, rxia\}@njust.edu.cn} \\
}

\begin{document}

\maketitle

\begin{abstract}
%Multimodal Aspect-Based Sentiment Analysis (MABSA) task gets increasing attention recently which has three subtasks: Multimodal Aspect Term Extraction, Multimodal Aspect-oriented Sentiment Classification and Joint Multimodal Aspect-Sentiment Analysis. However, previous approaches either use separately pre-trained visual and textual features which ignore the alignment between modalities or use vision-language models pre-trained only by general pre-training tasks which have weak ability to understand aspect-related objective expressions and opinion-related subjective expressions. 
As an important task in sentiment analysis, Multimodal Aspect-Based Sentiment Analysis (MABSA) has attracted increasing attention in recent years.
However, previous approaches either (i) use separately pre-trained visual and textual models, which ignore the cross-modal alignment or (ii) use vision-language models pre-trained with general pre-training tasks, which are inadequate to identify fine-grained aspects, opinions, and their alignments across modalities.
To tackle these limitations, we propose a task-specific Vision-Language Pre-training framework for MABSA (VLP-MABSA), which is a unified multimodal encoder-decoder architecture for all the pre-training and downstream tasks.
We further design three types of task-specific pre-training tasks from the language, vision, and multi-modal modalities, respectively.
Experimental results show that our approach generally outperforms the state-of-the-art approaches on three MABSA subtasks.
Further analysis demonstrates the effectiveness of each pre-training task.
The source code is publicly released at 
\url{https://github.com/NUSTM/VLP-MABSA}.

\end{abstract}
% \begin{CJK}{UTF8}{gbsn}
\section{Introduction}
%Joint Multi-modal Aspect-Sentiment Analysis (JMASA)Ju et al.( 2021)是Multimodal Sentiment Analysis领域最近提出的非常重要的一个任务.该任务旨在利用image-text pair信息,进行文本中aspect terms的抽取并classify相应aspect的sentiment.图1展示的是JMASA任务的示例，specifically,对于图1中左部分，给定上方的image-text pair,任务期望预测出文本中所有的aspect-sentiment pair, (Sergio Ramos,Positive) and (UCL,Neutral).如今社交媒体上拥有大量的多模态posts,JMASA任务可以很好地帮助我们分析推文中用户对特定entity的opinion.
Recent years have witnessed increasing attention on the Multimodal Aspect-Based Sentiment Analysis (MABSA) task\footnote{The MABSA task is also known as Target-Oriented Multimodal Sentiment Analysis or Entity-Based Multimodal Sentiment Analysis in the literature.}.
%In this paper, ``Aspect'' primarily refers to named entities such as Person, Location, and Organization.}.
Previous research mostly focused on its two subtasks, including Multimodal Aspect Term Extraction (MATE) and Multimodal Aspect-oriented Sentiment Classification (MASC).
Given a text-image pair as input, MATE aims to extract all the aspect terms mentioned in the text~\cite{zhang2018adaptive,lu:acl2018,wu2020multimodal,wu:2020mm,zhang2021multi}, whereas MASC aims to classify the sentiment towards each extracted aspect term~\cite{xu2019multi,yu2019adapting,khan2021exploiting}.
As the two subtasks are closely related to each other, \citet{ju2021emnlp} recently introduced the Joint Multimodal Aspect-Sentiment Analysis (JMASA) task, aiming to jointly extract the aspect terms and their corresponding sentiments.
%, which was recently proposed by.
%Given a text-image pair as input, the goal of MABSA is to extract all the aspect terms mentioned in the text and classify the sentiment towards each aspect term.
%As an important task in sentiment analysis, various kinds of approaches have been proposed for several MABSA subtasks, e.g., Multimodal Aspect Term Extraction~\cite{zhang2018adaptive,yu2020improving,wu2020multimodal,wu:2020mm}, Multimodal Aspect-oriented Sentiment Classification~\cite{xu2019multi,khan2021exploiting}, and Joint Multimodal Aspect-Sentiment Analysis~\cite{ju2021emnlp}.
%Take Joint Multimodal Aspect-Sentiment Analysis as an example. 
For example, given the text-image pair in Table.~\ref{Fig.motivation}, the goal of JMASA is to identify all the aspect-sentiment pairs, i.e., (\textit{Sergio Ramos}, \textit{Positive}) and (\textit{UCL}, \textit{Neutral}).

%\begin{figure}[!t] %H为当前位置，!htb为忽略美学标准，htbp为浮动图形
%\centering %图片居中
%\setlength{\abovecaptionskip}{0.1cm}
%\hbox{\includegraphics[scale=0.50]{fig/example2.pdf}} %插入图片，[]中设置图片大小，{}中是图片文件名
%\caption{An example of MABSA}
%label{Fig.motivation} %用于文内引用的标签
%\end{figure}

\begin{table}[!t]
\centering
\footnotesize
% \scriptsize
\setlength{\belowcaptionskip}{-1em}
\begin{tabular}{p{0.8cm}p{5cm}}
\toprule
Image & \begin{minipage}{0.1\textwidth}
    \hbox{\hspace{-0.2em} \includegraphics[width=50mm, height=30.0mm]{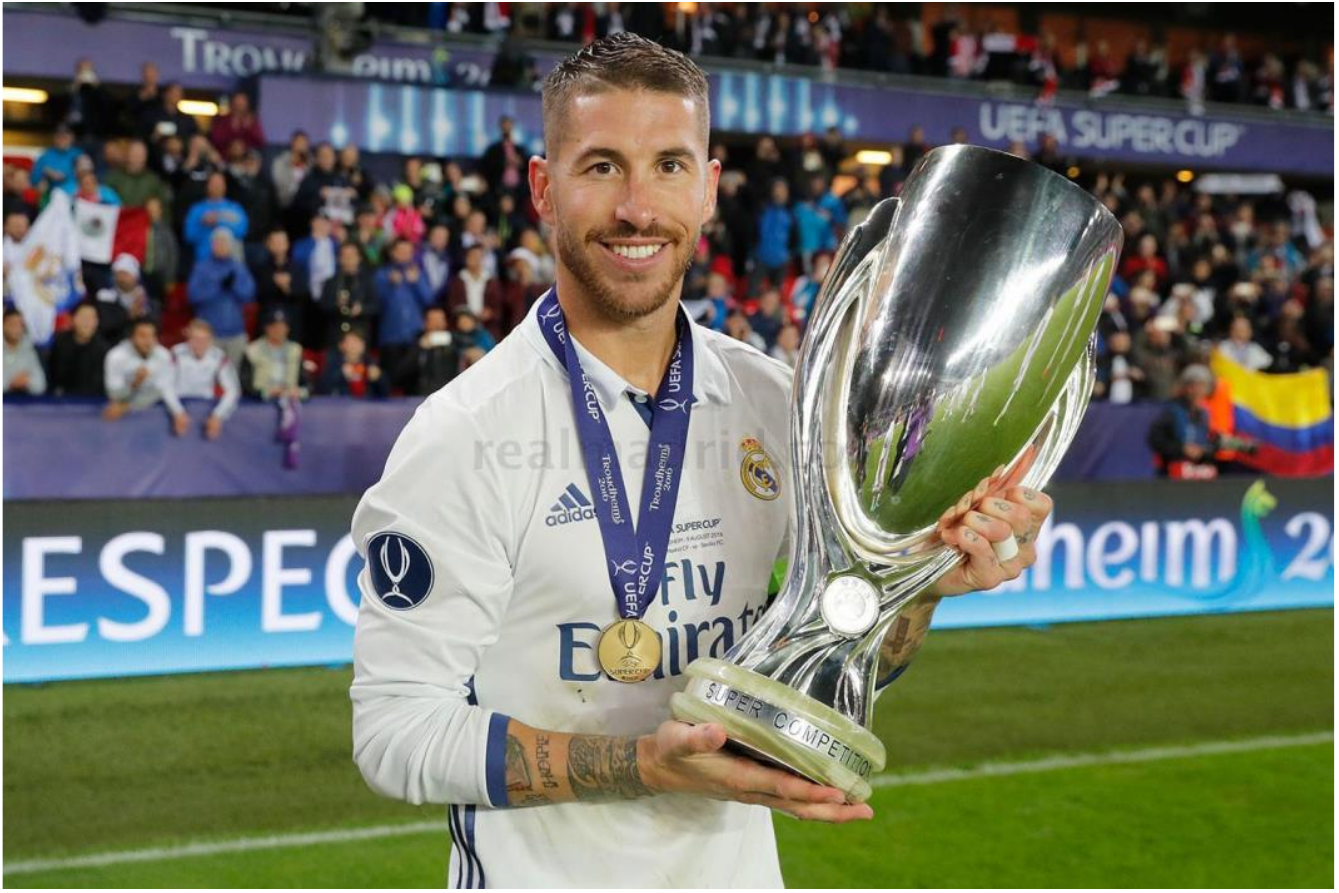}}
  \end{minipage}\vspace{0.3em}\\
%\midrule
\multirow{2}{*}{Text} &  Sergio Ramos chosen as the best player of UCL final\\
\midrule%\midrule
\multirow{2}{*}{Output} & (Sergio Ramos, Positive)\\
 & (UCL, Neutral)\\
\bottomrule
\end{tabular}
\caption{An example of the MABSA task}
\label{Fig.motivation}
\end{table}

Most of the aforementioned studies to MABSA primarily focused on employing pre-trained unimodal models (e.g., BERT for text and ResNet for image) to obtain textual and visual features respectively.
%With the success of applying pre-trained models in various downstream tasks, most recent MABSA studies follow this trend, and they can be roughly grouped into two categories.
%One line of work utilizes pre-trained models trained on single-modal data (e.g., BERT~\cite{devlin2018bert} for text and ResNet~\cite{he:cvpr2016} for images) to obtain the text and the image representations respectively~\cite{yu2019adapting,zhang2021multi}.
The separate pre-training of visual and textual features ignores the alignment between text and image.
It is therefore crucial to perform vision-language pre-training to capture such cross-modal alignment.
However, for the MABSA task, the studies on vision-language pre-training are still lacking.
%One attractive solution to build an alignment between visual and textual features is to perform vision-language pre-training~\cite{su2019vl,lu2019vilbert}, which has been shown to benefit many vision-language understanding tasks such as visual question answering and visual grounding~\cite{chen2020uniter}.
%For the tasks of MABSA, to the best of our knowledge, only very few existing studies explored the usefulness of visual-language pre-training~\cite{sun2020riva}.

To the best of our knowledge, there are very few studies focusing on vision-language pre-training for one of the MABSA subtasks, i.e., MATE~\cite{sun2020riva,sun2021rpbert}.
One major drawback of these studies is that they mainly employ general vision-language understanding tasks (e.g., text-image matching and masked language modeling) to capture text-image alignments. 
Such general pre-training is inadequate to identify fine-grained aspects, opinions, and their alignments across the language and vision modalities.
Therefore, it is important to design task-specific vision-language pre-training, to model aspects, opinions, and their alignments for the MABSA task.

\begin{figure*}[!tp] %H为当前位置，!htb为忽略美学标准，htbp为浮动图形
\centering %图片居中
\setlength{\abovecaptionskip}{0.2cm}
\setlength{\belowcaptionskip}{-0.3cm}
\hbox{\hspace{-0.3em}\includegraphics[scale=0.487]{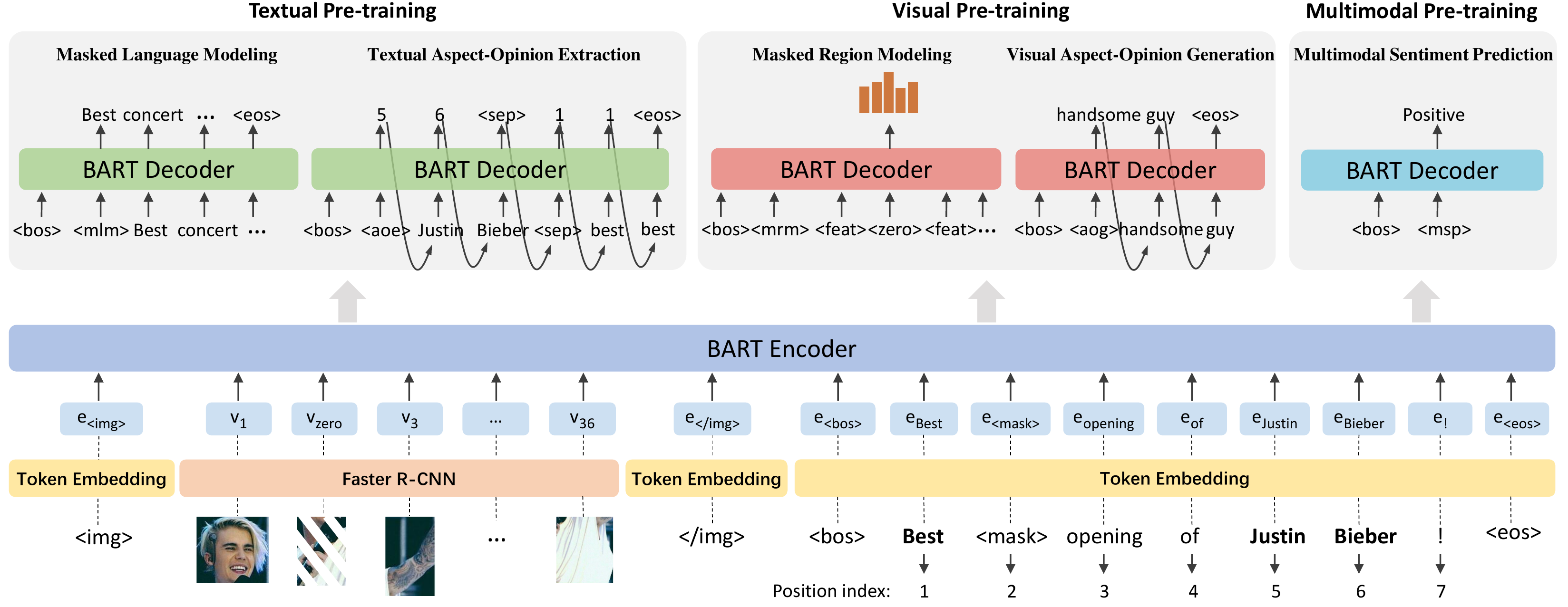}} %插入图片，[]中设置图片大小，{}中是图片文件名
\caption{Overview of our Vision-Language Pre-Training framework for MABSA}
\label{Fig.overview} %用于文内引用的标签
\end{figure*}

To address this issue, in this paper, we propose a task-specific Vision-Language Pre-training framework for Multimodal Aspect-Based Sentiment Analysis.
Specifically, inspired by the recent success of BART-based generative models in text-based ABSA~\cite{yan2021unified}, we first construct a generative multimodal architecture based on BART~\cite{lewis2020bart}, for both vision-language pre-training and the downstream MABSA tasks.
%by formulating all the pre-training and MABSA tasks as generation problems.
We then propose three types of vision-language pre-training tasks, including Masked Language Modeling (MLM) and Textual Aspect-Opinion Extraction (AOE) from the language modality, Masked Region Modeling (MRM) and Visual Aspect-Opinion Generation (AOG) from the vision modality, and Multimodal Sentiment Prediction (MSP) across two modalities.
Figure~\ref{Fig.overview} illustrates the whole framework of our proposed pre-training approach.
Compared with general pre-training methods, our task-specific pre-training approach incorporates multimodal aspect, opinion, and sentiment supervision, which guides pre-trained models to capture important objective and subjective information for the MABSA task.
%
%our designed task-specific pre-training tasks capture the objective and subjective information from image and text which is beneficial to the downstream task MABSA. 

%For textual modality, apart from , we introduce an Aspect and Opinion Co-Extraction task to identify objective and subjective expressions in text.
%Similarly, for visual modality, we employ the general  task, and also design an Adjective-Noun Generation task to better capture subjective and objective information in images.
%Lastly, we incorporate a Multimodal Sentiment Prediction task to strengthen model's ability in fusing textual and visual subjective information for sentiment classification.
%预训练数据和监督信号，每个两三行，和预训练的术语结合起来。三类、五种总的描述，名字给出来不一定要介绍。

%我们设计了三类预训练任务,分别是基于text,基于image,multimodol.text的预训练包括Masked Language Model(MLM),Aspect  and Opinion Co-Extraction(AOCE)来挖掘text中主客观信息;image的预训练任务包括Masked Region Model(MRM), Adjective and Noun Generation(ANG)来挖掘image中主客观信息;multimodal task Sentiment Prediction(SP)来增强模型理解多模态sentiment的能力。

%under which all the pre-training tasks and MABSA tasks are formulated as generative problems

%Pre-training datasets \& Labels for each task ???
To evaluate the effectiveness of our pre-training approach, we adopt MVSA-Multi, a widely-used Multimodal Twitter dataset for coarse-grained text-image sentiment analysis~\cite{niu2016sentiment}, as our pre-training dataset.
%For each multimodal tweet in this dataset, 
We then employ several representative pre-trained models and rule-based methods to obtain the aspect and opinion supervision for our AOE and AOG tasks.
%, since the two benchmark MABSA datasets~\cite{ju2021emnlp} are collected from Twitter.
%For each multimodal tweet in this dataset, we utilize a well-known entity extraction toolkit and a sentiment lexicon to obtain the textual aspect and opinion supervision for our AOE task, and employ a pre-trained adjective-noun pair extractor to obtain the visual aspect and opinion supervision for our AOG task.
As the dataset provides sentiment labels for each multimodal tweet, we adopt them as the supervision for our MSP task.
%For each multimodal tweet in this dataset, we propose some strategies to obtain pseudo labels as the supervision signal for our AOE and AOG task.

%实验证明我们设计的预训练任务十分有效，使用完整的预训练在所有下游任务MABSA,MATE,MASC上generally outperform other approaches. Futher more,we通过supervised and weakly supervised training证明了各个预训练任务的有效性.
%我们还测试了只使用少量样本进行下游训练，结果显示预训练可以带来巨大提升,F1 score在两个数据集上分别提升14个点和5个点。

Our contributions in this work are as follows:
\vspace{-3pt}
\begin{itemize}[leftmargin=0.15in]
\setlength\itemsep{-0.2em}
\item
We introduce a task-specific Vision-Language Pre-training framework for MABSA named VLP-MABSA, which is a unified multimodal encoder-decoder architecture for all the pre-training and downstream tasks.
%(which is easy-to-implement and can be built on the pre-trained models, such as BART.)
\item
Apart from the general MLM and MRM tasks, we further introduce three task-specific pre-training tasks, including Textual Aspect-Opinion Extraction, Visual Aspect-Opinion Generation, and Multimodal Sentiment Prediction, to identify fine-grained aspect, opinions, and their cross-modal alignments.

\item
Experiments on three MABSA subtasks show that our pre-training approach generally obtains significant performance gains over the state-of-the-art methods.
Further analysis on supervised and weakly-supervised settings demonstrates the effectiveness of each pre-training task.
%We further design three types of pre-training tasks from the unimodality and multimodality, respectively.
%我们设计了三大类预训练任务,包括text-based,image based,multimodal.text-based包括MLM,AOCE,image-based包括MRM,ANG,multimodal包括SP,实验结果显示所有的预训练任务均effective。

\end{itemize}
% \end{CJK}

%contribution加不加都可以, 在future work提一下需要用大规模数据进行预训练

\begin{CJK}{UTF8}{gbsn}
\section{Related Work}

\textbf{Vision-Language Pre-training.}
Inspired by the success of pre-trained language models like BERT~\cite{devlin2019bert}, many multimodal pre-training models have been proposed~\cite{chen2020uniter,yu2021ernie,zhang2021vinvl} to perform many vision-language tasks which achieve fantastic success. Correspondingly, many general pre-training tasks are proposed, such as Masked Language Modeling (MLM), Masked Region Modeling (MRM) and Image-Text Matching (ITM)~\cite{chen2020uniter,yu2021ernie}. Besides, in order to make the pre-trained models better understand downstream tasks, researchers also design task-specific pre-training 
models for different downstream tasks~\cite{hao2020towards, xing2021km}.
%起初主要集中于understanding task，后来为了解决vision-language generation task，如Image Caption(You et al., 2016)[8]，VCG(Park et al., 2020)[10]等，使得模型拥有优秀的generation能力，许多基于transformer encoder-decoder框架的预训练模型被提出，并且为了使得模型更好地理解特定的任务，还设计了针对任务的预训练策略，例如Action Prediction(Hao et al., 2020)[11]和KCG(Xing et al., 2021)[9]任务。
In our work, apart from the popular general pre-training tasks, we also design three kinds of task-specific pre-training tasks for the MABSA task.

\textbf{Text-based Joint Aspect-Sentiment Analysis (JASA).}
%Aspect extraction and aspect-level sentiment classification是Aspect Based Sentiment Analysis中两个重要的子任务，which aim to extract aspect terms and identify the sentiment orientations towards them. In the literature, aspect extraction (Qiu et al., 2011; Liu et al., 2015; Poria et al., 2016; Wang et al., 2016a, 2017; Li et al., 2018a; Xu et al., 2018) and aspect-level sentiment classification (Dong et al., 2014; Tang et al., 2016; Wang et al., 2016b; Ma et al., 2017; Wang et al., 2018; Li et al., 2019c, Sundararaman et al., 2020; Ji et al., 2020; Liang et al., 2020b,a) have been extensively studied.
JASA aims to extract aspect terms in the text and predict their sentiment polarities. 
Many approaches have been proposed including pipeline approaches~\cite{zhang2015neural,hu2019open}, multi-task learning approaches~\cite{acl/HeLND19,hu2019open} and collapsed label-based approaches~\cite{li2019unified,hu2019open,coling/ChenTS20}.
Recently, \citet{yan2021unified} proposed a unified generative framework which achieves highly competitive performance on several benchmark datasets for JASA.
%In this work, our  their framework

\textbf{Multimodal Sentiment Analysis.}
Multimodal Sentiment Analysis (MSA) in social media posts is an important direction of sentiment analysis. 
Many neural network approaches have been proposed to perform the coarse-grained MSA in the literature, which aim to detect the overall sentiment of each input social post~\cite{you2015joint,you2016cross,luo2017social,xu2018co,yang2021multimodal}. 
Different from these studies, our work focuses on the fine-grained MABSA task, which aims to identify the sentiments towards all the aspects mentioned in each input social post.

\textbf{Multimodal  Aspect-Based  Sentiment  Analysis.}
As an important sentiment analysis task, many approaches have been approached to tackle the three subtasks of MABSA, including Multimodal Aspect Term Extraction~\cite{zhang2018adaptive,yu2020improving,wu2020multimodal,wu:2020mm,sun2020riva,zhang2021multi}, Multimodal Aspect Sentiment Classification~\cite{xu2019multi,yu2019entity,yang2021fine,khan2021exploiting} and Joint Multimodal Aspect-Sentiment Analysis~\cite{ju2021emnlp}.
In this work, we aim to propose a general pre-training framework to improve the performance of all the three subtasks.

%MABSA has witnessed increasing attention in recent years. The studies towards MABSA can divide into three subtasks: MATE, MASC and JMASA. 

%In the literature，许多研究集中于JMASA的两个子任务Multimodal Aspect Terms Extraction(MATE) and Multimodal Aspect Sentiment Classification (MASC).这方面许多工作研究如何将image和text合适地融合，对于MATE,如使用RNN~\cite{zhang2018adaptive},使用Transformer~\cite{yu2020improving},使用GNN~\cite{zhang2021multi};对于MASC,如使用BERT~\cite{yu2019adapting},使用Memory Network~\cite{xu2019multi},将image转化为caption~\cite{khan2021exploiting}.然而在实际的应用场景中，text中的aspect term并不会给定，所以Ju et al.(2021)提出了Joint Multi-modal Aspect-Sentiment Analysis(JMASA)任务，同时进行aspect terms的抽取和aspect-targeted sentiment classification，并提出利用辅助任务来控制图像信息。

%Multimodal  Aspect-Based  Sentiment  Analysis(MABSA)任务is still under-explored.研究主要集中于其两个子任务Multimodal Aspect Terms Extraction(MATE)和Multimodal Aspect-oriented Sentiment Classification(MASC)。

%对于MATE，许多在named entity recognition任务上的工作，利用整个图像的编码信息来增强text的表示能力，比如(Moon et al., 2018; Zhang et al., 2018, Yu et al., 2020b ,Zhang et al., 2021b)。同时，一些研究提出利用细粒度的图像信息通过目标检测，如(Wu et al., 2020a,b)；

%对于MASC，许多工作针对如何融合多模态特征(Xu et al.,2019,;Yu and Jiang,2019;Yu et al.2020a)以及丰富文本表示(Khan and Fu,2021)提出了很多方法.

%Recently,\citet{ju2021emnlp}提出此任务，jointly perform MATE and MASC,并且提出了multi-modal joint learning approach with auxiliary cross-modal relation detection来解决 。
%本文解决这个任务的方法是首先利用额外数据进行预训练，然后在下游上进行finetuning。

\end{CJK}
% \begin{CJK}{UTF8}{gbsn}

\section{Methodology}
%在这个章节，我们讲述了我们解决MABSA任务的方法。3.1讲述了我们的模型结构，3.2讲述了我们的预训练任务, 3.3讲述了下游任务。
% In this section, we describe our method for MABSA. Section 3.1 
Figure~\ref{Fig.overview} shows the overview of our model architecture. The backbone of our model is BART~\cite{lewis2020bart}, which is a denoising autoencoder for sequence-to-sequence models. 
We extend BART to encode both textual and visual inputs, and decode pre-training and downstream tasks from different modalities.
In the following subsections, we first introduce our feature extractor, and then illustrate the encoder and decoder of our model, followed by describing the details of three types of pre-training tasks and downstream MABSA tasks.
%合适的地方引用一下KM-BART
\subsection{Feature Extractor}
% In this section, we introduce our feature extractor of two modalities.
\textbf{Image Representation.} Following many existing Vision-Language pre-training models~\cite{chen2020uniter, yu2021ernie}, we employ Faster R-CNN~\cite{anderson2018bottom} to extract visual features. 
Specifically, we adopt Faster R-CNN to extract all the candidate regions from an input image. 
We then only retain 36 regions with the highest confidence. 
Meanwhile, we also keep the semantic class distribution of each region, which will be used for the Masked Region Modeling task. 
For the retained regions, we use mean-pooled convolutional features processed by Faster R-CNN as our visual features. 
Let us use $R=\{\mathit{r}_1,..., \mathit{r}_{36}\}$ to denote the visual features, where $\mathit{r_i}\in\mathbb{R}^{2048}$ refers to the visual feature of the $i$-th region. 
To be consistent with the text representation, we adopt a linear transformation layer to project visual features to $d$-dimensional vectors, denoted by $\mathit{V}\in\mathbb{R}^{d\times36}$.

% \noindent
\textbf{Text Representation.} 
For text input, we first tokenize the text and then feed tokens to the embedding matrix. 
The embeddings of text tokens are used as text features. 
Let us use $E=\{e_1,...,e_T\}$ to denote the token indexes of text inputs where $T$ denotes the length of the input text, and $\mathbf{W}=\{\mathbf{w}_1,...,\mathbf{w}_T\}$ to denote the embeddings of tokens.

%Unlike the original BART only input text, the input of our model is multi-modal.
%\subsection{Encoder \& Decoder}
\subsection{BART-based Generative Framework}
We employ a BART-based generative framework for both vision-language pre-training and downstream MABSA tasks. 

\textbf{Encoder.}
The encoder of our model is a multi-layer bidirectional Transformer. 
As shown in Figure~\ref{Fig.overview}, %which is also a visual-language pre-training framework based on BART,
to distinguish inputs of different modalities, we follow~\citet{xing2021km} by using $\langle\textit{img}\rangle$ and $\langle\textit{/img}\rangle$ to indicate the start and the end of visual features, and $\langle \textit{bos}\rangle$ and $\langle \textit{eos}\rangle$ to indicate the textual input. 
%We denote text inputs by$W=\{\mathit{w}_1,...,\mathit{w_L}\}$,$\mathit{L}$是文本输入的token数量,包括word和标识符,$w_i\in\mathbb{R}^{768}$表示第i个token的embedding, 768是embedding的维度.
In the following part of the paper, we denote the concatenated multimodal input by $X$. %$X=\{e_{<img>},...,v_i,...,e_{</img>},e_{<bos>},e_1,...,e_T,e_{<eos>}\}$

%\noindent
\textbf{Decoder.}
%在不同的预训练任务上，我们都用了同样的decoder结构。
The decoder of our model is also a multi-layer Transformer. The difference is that the decoder is unidirectional %autoregressive 
when generating outputs, while the encoder is bidirectional. 
Since all pre-training tasks share the same decoder, we insert two special tokens at the beginning of the inputs of the decoder to indicate different pre-training tasks. 
Following \citet{yan2021unified}, we insert a special token $\langle\textit{bos}\rangle$ to indicate the beginning of generation, and then insert a task-specific special token to indicate the task type. 
Specifically, the special tokens for Masked Language Modeling, Textual Aspect-Opinion Extraction, Masked Region Modeling, Visual Aspect-Opinion Generation, and Multimodal Sentiment Prediction are $\langle\textit{bos}\rangle\langle\textit{mlm}\rangle$, $\langle\textit{bos}\rangle\langle\textit{aoe}\rangle$, $\langle\textit{bos}\rangle\langle\textit{mrm}\rangle$, $\langle\textit{bos}\rangle\langle\textit{aog}\rangle$, and $\langle\textit{bos}\rangle\langle\textit{msp}\rangle$, respectively.
% \textit{<bos><mlm>}, \textit{<bos><aoe>}, \textit{<bos><mrm>}, \textit{<bos><aog>}, and \textit{<bos><msp>}, respectively.
%For Masked Language Modeling, the special tokens are \textit{<bos><mlm>}, for Textual Aspect-Opinion Extraction we add \textit{<bos><aoe>}, for Masked Region Modeling we add \textit{<bos><mrm>}, for Visual Aspect-Opinion Generation we add \textit{<bos><aog>}, for Sentiment Prediction, we add \textit{<bos><sp>}.

\subsection{Pre-training Tasks}
\iffalse
\begin{table*}[!tp]
\centering
\scriptsize
\setlength{\abovecaptionskip}{0.30cm}
\begin{tabular}{cccccccc}
\toprule
& Aspect-all & Aspect-avg & Opinion-all & Opinion-avg & word-all & word-avg & image-text pair \\
\midrule
Positive & 10593 & 0.899 & 22752 & 1.911 & 215044 & 18.066 & 11903\\
Neutral & 3756 & 0.914 & 7567 & 1.842 & 74456 & 18.129 & 4107\\
Negtive & 1016 &0.677 & 2956 & 1.977 & 25211 & 16.807 & 1500\\
%Totoal & 15365 &
\bottomrule
\end{tabular}
\caption{\footnotesize Statistics of MVSA-Multi Dataset}
\label{tab:mvsa}
\end{table*}
\fi

\begin{table}[!tp]
\centering
\scriptsize
\setlength{\belowcaptionskip}{-1.3em}
\begin{tabular}{ccccc}
\toprule
Sentiment &  \#Image-Text Pairs  &  \#Aspects & \#Opinions & \#Words\\
\midrule
Positive & 11903 & 10593 & 22752 & 215044  \\
Neutral  & 4107  & 3756  & 7567  & 74456   \\
Negative  & 1500  & 1016  & 2956  & 25211   \\
%Totoal & 15365 &
\bottomrule
\end{tabular}
\caption{\footnotesize The statistics of the MVSA-Multi Dataset. \#Apects and \#Opinions are the number of aspect terms and opinion terms we extract from the dataset by the rule-based methods introduced in Section~\ref{text_pretrain}. }
\label{tab:mvsa}
\end{table}

% \textbf{Pre-training Dataset.}
The dataset we use for pre-training is MVSA-Multi~\cite{niu2016sentiment}, which is widely used in Multimodal Twitter Sentiment Analysis~\cite{yadav2020sentiment, yang2021multimodal}. 
This dataset provides image-text input pairs and coarse-grained sentiments of image-text pairs. Statistics of the dataset are given in Table~\ref{tab:mvsa}.
%提一下是粗粒度的标注，表格设计和预训练三个任务结合，可分为三个子表

With the dataset, we design three types of pre-training tasks, including textual, visual, and multimodal pre-training as follows. 
%In the following, we will introduce these pre-training methods in detail.

\subsubsection{Textual Pre-training}
\label{text_pretrain}
Textual Pre-training contains two tasks: a general Masked Language Modeling task to build alignment between textual and visual features and a task-specific Textual Aspect-Opinion Extraction task to extract aspects and opinions from text.  

%\noindent
\textbf{Masked Language Modeling (MLM).} 
In the MLM pre-training task, we use the same strategy as BERT~\cite{devlin2019bert} by randomly masking the input text tokens with a probability of 15\%.
The goal of the MLM task is to generate the original text based on the image and the masked text, 
%We use $h_i^d$ to denote the current step output of the decoder, and use E to denote the set of all token indexes of BART tokenizer. $Embed$ indicates the embedding matrix. 
% So the distribution over all tokens is:
% \begin{equation}
% \begin{aligned}
% P(E|X)= Softmax(Embed \cdot h_i^d)
% \end{aligned}
% \end{equation}
and thus the loss function of the MLM task is:
\begin{equation}
\begin{aligned}
\mathcal{L}_{MLM}=-\mathbb{E}_{X\sim D} \sum_{i=1}^T\log\mathit{P(e_i|e_{<i},\tilde{X})},
\end{aligned}
\end{equation}
where $e_{i}$ and $\tilde{X}$ denote the $i^{th}$ token of the input text and the masked multimodal input, respectively. T is the length of input text.

%\noindent
\textbf{Textual Aspect-Opinion Extraction (AOE).} 
The AOE task aims to extract aspect and opinion terms from the text. Since the MVSA-Multi dataset does not provide annotations for aspect and opinion terms, we resort to a pre-trained model for aspect extraction and a rule-based method for opinion extraction. 
Specifically, for aspect extraction, we employ the pre-trained model from a well-known Named Entity Recognition (NER) tool for tweets~\cite{Ritter11} to perform NER on each tweet in the dataset, and regard the recognized entities as aspect terms. 
For opinion extraction, we utilize a widely-used sentiment lexicon named SentiWordNet~\cite{esuli2006sentiwordnet} to obtain the dictionary of opinion words. 
Given each tweet, if its sub-sequences (i.e., words or phrases) match the words in the dictionary, we treat them as opinion terms. These extracted aspect and opinion terms are used as the supervision signal of our AOE task. 
%These extracted aspect and opinion terms are used as the supervision signal of our AOE task. 

%特别像的特别要引用别人的工作，如下面的公式
%We use $a$ and $o$ to denote aspect and opinion terms, and use superscript $^s,^e$ to denote the start index and the end index of a term. 
With the textual aspect-opinion supervision, we follow \citet{yan2021unified} by formulating the AOE task as an index generation task. 
Given the input text as the source sequence, the goal is to generate a target index sequence which consists of the start and end indexes of all aspect and opinion terms. 
Let us use $Y\!=\![a^s_1,a^e_1,...,a^s_M,a^e_M,\langle\textit{sep}\rangle,o^s_1,o^e_1,...,o^s_N,o^e_N,\langle\textit{eos}\rangle]$ to denote the target index sequence, where $M$ and $N$ are the number of aspect terms and opinion terms, $a^s$,$a^e$ and $o^s$,$o^e$ indicate the start and end indexes of an aspect term and an opinion term respectively, $\langle\textit{sep}\rangle$ is used to separate aspect terms and opinion terms, and $\langle\textit{eos}\rangle$ informs the end of extraction.
For example, as shown in Figure~\ref{Fig.overview}, the extracted aspect and opinion terms are \textit{Justin Bieber} and \textit{best} respectively, and the target sequence is $Y\!=\![5,6,\langle\textit{sep}\rangle,1,1,\langle\textit{eos}\rangle]$.
For $y_t$ in the target sequence $Y$, it is either a position index or a special token (e.g., $\langle\textit{sep}\rangle$). We use $\mathit{C=[\langle\textit{sep}\rangle, \langle\textit{eos}\rangle]}$ to denote the set of special tokens, and $\mathbf{C}^d$ as their embeddings.

We assume that $\mathbf{H}^e$ denotes the encoder output of the concatenated multimodal input, $\mathbf{H}^e_{T}$ denotes the textual part of $\mathbf{H}^e$, and $\mathbf{H}^e_{V}$ denotes the visual part of $\mathbf{H}^e$. %the encoder output of the textual input.
The decoder takes the multimodal encoder output $\mathbf{H}^e$ and the previous decoder output $Y_{<t}$ as inputs, and predicts the token probability distribution $P(\mathit{y_t})$ as follows:
%\begin{align}
\begin{align}
\label{eqn2}
\mathbf{h}_t^d&=\text{Decoder}(\mathbf{H}^e;Y_{<t}),\\
\bar{\mathbf{H}}_{T}^e&=(\mathbf{W}+\mathbf{H}^e_T)/2,\\
\label{eqn4}
P(\mathit{y_t})&=\text{Softmax}([\bar{\mathbf{H}}_{T}^e;\mathbf{C}^d]\mathbf{h}_t^d),
% \end{equation}
\end{align}
% \begin{align}
% &E^e=BARTTokenEmbed(W)\\
% &\overline{H}^e=(H^e+E^e)/2\\
% &C_d=BARTTokenEmbed(C)\\
% &P_t=Softmax([\overline{H}^e;C_d]h_t^d)
% \end{align}
% \begin{equation}
% \setlength{\abovedisplayskip}{3pt} %%% 3pt 
% \setlength{\belowdisplayskip}{3pt}
% \begin{aligned}
% \end{aligned}
% \end{equation}
%where $C_d$ denotes the embeddings of functional tokens. And 
where $\mathbf{W}$ denotes the embeddings of input tokens.
The loss function of the AOE task is as follows:
\begin{equation}
\begin{aligned}
\mathcal{L}_{AOE}=-\mathbb{E}_{X\sim D}\sum_{t=1}^{O}\log P(\mathit{y_t|Y_{<t},X}),
\end{aligned}
\end{equation}
where $O=2M+2N+2$ is the length of $Y$ and $X$ denotes the multimodal input.

\subsubsection{Visual Pre-training}
Visual Pre-training contains two tasks: a general Masked Region Modeling task and a task-specific Visual Aspect-Opinion Generation task to capture subjective and objective information in the image.

%\noindent
\textbf{Masked Region Modeling (MRM).} 
Following \citet{xing2021km}, our MRM task aims to predict the semantic class distribution of the masked region.
%, we formulate the MRM task as a masked sequence-to-sequence 
As shown in Figure~\ref{Fig.overview}, 
for the input of the encoder, we randomly mask image regions with a probability of 15\%, which are replaced with zero vectors. 
For the input of the decoder, we first add two special tokens $\langle\textit{bos}\rangle\langle\textit{mrm}\rangle$, and then represent each masked region with $\langle\textit{zero}\rangle$ and each remaining region with $\langle\textit{feat}\rangle$.
After feeding the input to the decoder, an MLP classifier is stacked over the output of each $\langle\textit{zero}\rangle$ to predict the semantic class distribution. 
Let us use $p(v_z)$ to denote the predicted class distribution of the $z$-th masked region, and $q(v_z)$ to denote the class distribution detected by Faster R-CNN.
% \begin{equation}
% \begin{aligned}
% \mathit{q(f_z)}=Softmax(MLP(H_z^d))
% \end{aligned}
% \end{equation}
%1$<=z<=Z$,Z为mask的region数量.
The loss function for MRM is to minimize the KL divergence of the two class distributions:
\begin{equation}
\begin{aligned}
\mathcal{L}_{MRM}=\mathbb{E}_{X\sim D}\sum_{z=1}^ZD_{KL}(\mathit{q(v_z)||p(v_z)}),
\end{aligned}
\end{equation}
where $Z$ is the number of masked regions.
%\subsubsection{Adjective Noun Pair Distribution Prediction(ADP)}
%Adjective Noun Pairs(ANPs)~\cite{borth2013large},combine the sentimental strength of adjectives and detectability of nouns, such as “happy dog” and “beautiful sky”.这是很好的衔接low-level的图像特征与high-level的sentiment的方式。Adjective Noun Pair Distribution Prediction这个任务旨在通过模型预测distribution over事先定义好的2089类ANP。我们首先使用一个ANP detector DeepSentiBank~\cite{chen2014deepsentibank},对数据中每张图片计算其ANP distribution。然后在decoder端输入任务标志token <bos><ANP>,接着使用<ANP>在decoder端的输出特征，送入一个MLP分类器，得到我们预测的ANP distribution。为The loss function is to minimize the KL divergence of the predict distribution and the distribution detected by DeepSentiBank.我们用p,q分别表示detector得到的ANP distribution和我们模型预测得到ANP distribution，则损失函数为
%$$\mathcal{L}_{ADP}(X)=KL(p||q)$$

%\noindent
\textbf{Visual Aspect-Opinion Generation (AOG).} 
The AOG task aims to generate the aspect-opinion pair detected from the input image. 
In the field of Computer Vision, \citet{borth2013large} proposed to detect the visual sentiment concept, i.e., Adjective-Noun Pair (ANP) such as \textit{smiling man} and \textit{beautiful landscape} in the image.
Since the nouns and adjectives of ANP respectively capture the fine-grained aspects and opinions in the image, we regard ANPs as visual aspect-opinion pairs.
%detect the generic visual concepts (i.e., nouns such as \textit{woman}, \textit{guy}) and identify their correlated visual sentiments (i.e., adjectives such as \textit{smiling}, \textit{handsome}).
%Adjective-Noun Pairs (ANP) are a popular mid-level semantic construct for capturing fine-grained aspect and opinion in an image via visually detectable concepts such as “cute dog" or “beautiful landscape".
% In AOG task, we use a ANP detector\footnote{https://github.com/stephen-pilli/DeepSentiBank} DeepSentiBank~\cite{chen2014deepsentibank} to detect ANP distribution which has 2089 classes for each image in the dataset. After that, the one with the highest probability after softmax is selected as our target ANP. Specifically, as shown in Fig.~\ref{Fig.overview}, the ANP with the highest probability we detect from the input image is "handsome guy". "handsome" gives the clue of opinion and "guy" gives the clue of aspect. This can be considered as a good reflection of subjective and objective information extracted from image. We first input the task token <bos><aog> to the decoder, and then generate the target ANP in way of autoregressive.
In order to detect the ANP of each input image, we adopt a pre-trained ANP detector DeepSentiBank\footnote{https://github.com/stephen-pilli/DeepSentiBank}~\cite{chen2014deepsentibank} to predict the class distribution over 2089 pre-defined ANPs. 
The ANP with the highest probability is selected as the supervision signal of our AOG task.
%which is used as the supervision signal. 
%We conduct AOG task to generate the ANP through our BART-based generative framework. Specifically, 
For example, in Figure~\ref{Fig.overview}, the ANP detected from the input image is \textit{handsome guy}, and we regard it as the supervision. 

With the visual aspect-opinion supervision, we formulate the AOG task as a sequence generation task.
Specifically,  
%first two task tokens \textit{<bos><aog>} are first fed to the decoder, and then the aspect-opinion pairs are generated under the supervision in an auto-regressive manner.
%我们选取softmax后概率最大的作为我们的目标ANP. Specifically, as shown in fig1, the ANP we detect from the input image is "handsome guy". "handsome" gives the clue of opinion and "guy" gives the clue of aspect.这很好地反映了图像中地主客观信息。 AOG任务通过在decoder端输入任务token<bos><aog>, 采用自回归的形式,期望生成目标ANP. 我们使用$S=\{s_1,...,s_i,...,s_A\}$表示目标ANP的tokens,A表示ANP含有的token数量,则AOG任务的loss为
let us use $G=\{g_1,...,g_{|G|}\}$ to denote the tokens of the target ANP and $|G|$ to denote the number of ANP tokens.
The decoder then takes the multimodal encoder output $\mathbf{H}^e$ and the previous decoder output $G_{<i}$ as inputs, and predicts the token probability distribution $P(g_i)$:
\begin{align}
\mathbf{h}_i^d&=\text{Decoder}(\mathbf{H}^e;G_{<i}),\\
P(g_i)&=\text{Softmax}(\mathbf{E}^T \mathbf{h}_i^d),
\end{align}
where $\mathbf{E}$ denotes the embedding matrix of all tokens in the vocabulary.

The loss function of the AOG task is:
\begin{equation}
\begin{aligned}
\mathcal{L}_{AOG}=-\mathbb{E}_{X\sim D}\sum_{i=1}^{|G|}\log\mathit{P(g_i|g_{<i},X).}
\end{aligned}
\end{equation}

\subsubsection{Multimodal Pre-training}
%Multimodal Pre-training has one task named Sentiment Prediction. Since judging the sentiment only by unimodality is not accurate enough, this task combines multimodal information to predict the sentiment at the image-text pair level.
Multimodal Pre-training has one task named Multimodal Sentiment Prediction (MSP).  
Different from the aforementioned pre-training tasks whose supervision signals only come from one modality, the supervision signals for MSP come from multimodality, which can enhance models to identify the subjective information in both language and vision and capture their rich alignments.

%\noindent
\textbf{Multimodal Sentiment Prediction (MSP).}
%我们利用了MVSA-Multi数据集中image-text级别上sentiment的标注 which 很好地反映了image，text综合的情绪表达。对于Sentiment Prediction(SP)任务，我们在decoder端输入任务标识符<bos><sp>,对<sp>经过decoder的输出$h_{sp}^d$接一个MLP classifier进行sentiment的预测。我们使用S来表示sentiment类别集合,则the sentiment distribution is
As the MVSA-Multi dataset provides the coarse-grained sentiment labels for all the text-image pairs, we use the sentiment labels as supervision signals of our MSP task. 
Formally, we model the MSP task as a classification task, where we first feed the two special tokens $\langle\textit{bos}\rangle\langle\textit{msp}\rangle$ to the decoder and then predict the sentiment distribution $P(s)$ as follows:
%We first input the task token \textit{<bos><sp>} to the decoder and predict the distribution of sentiment as follows:
\begin{align}
\mathbf{h}_{msp}^d&=\text{Decoder}(\textbf{H}^e;\mathbf{E}_{msp}),\\
P(s)&=\text{Softmax}(\text{MLP}(\mathbf{h}_{msp}^d)),
\end{align}
where $\mathbf{E}_{msp}$ is the embeddings of two special tokens.

We use the cross-entropy loss for the MSP task:
\begin{equation}
\begin{aligned}
\mathcal{L}_{MSP}=-\mathbb{E}_{X\sim D}\log\mathit{P(s|X)},
\end{aligned}
\end{equation}
where $s$ is the golden sentiment annotated in dataset.
\subsubsection{Full Pre-training Loss}
% 为了将上述提到的所有loss合并运用到训练过程中，我们为每个loss设置了合适的权重然后将它们加权作为最后的loss,权重分别为$\mathit{W_{MLM},W_{AOCE},W_{MRM},W_{ANG},W_{SP}} \in \mathbb{R} $。最后模型的loss为
To optimize all the model parameters, we adopt the alternating 
optimization strategy to iteratively optimize our five pre-training tasks. 
The objective function is as follows:
%To combine all the pre-training tasks in the pre-training stage, we adopt the objective function as follows:
%we weight the losses described above by $W_{MLM}$, $W_{AOE}$, $W_{MRM}$, $W_{AOG}$, $W_{MSP}$ to make a balance. Our full pre-training loss is
% \begin{equation}
% \begin{aligned}
% \mathcal{L}=\mathcal{L}_{MLM} + \mathcal{L}_{AOE} + \mathcal{L}_{MRM}
% +\mathcal{L}_{AOG}
% +\mathcal{L}_{MSP}.
% \end{aligned}
% \end{equation}
\begin{equation}
\begin{aligned}
\mathcal{L}=&\lambda_1\mathcal{L}_{MLM} + \lambda_2\mathcal{L}_{AOE}+\lambda_3\mathcal{L}_{MRM}+\\
&
\lambda_4\mathcal{L}_{AOG}
+\lambda_5\mathcal{L}_{MSP}
\end{aligned}
\end{equation}
where $\lambda_1$, $\lambda_2$, $\lambda_3$, $\lambda_4$, and $\lambda_5$ are tradeoff hyper-parameters to control the contribution of each task.

\begin{figure}[!t] %H为当前位置，!htb为忽略美学标准，htbp为浮动图形
\centering %图片居中
\hbox{\hspace{-0.2cm}\includegraphics[scale=0.45]{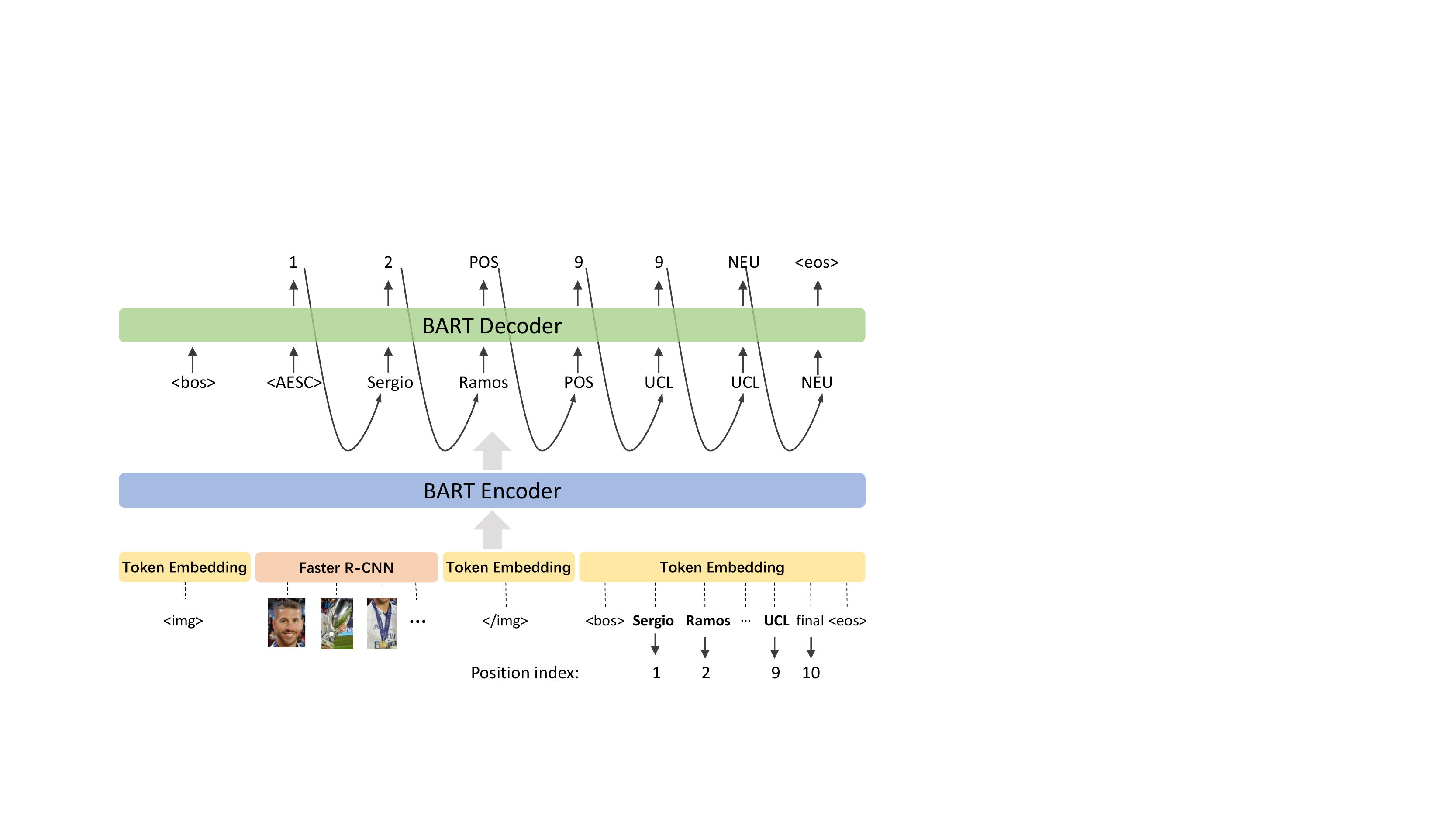}} %插入图片，[]中设置图片大小，{}中是图片文件名
\caption{An example of downstream task JMASA. $\langle\textit{AESC}\rangle$ informs the current task is JMASA.}
\label{Fig.downstream} %用于文内引用的标签
\end{figure}

\subsection{Downstream Tasks}
%提一下同构，比较好的用到下游任务
We consider all the three subtasks in MABSA as our downstream tasks, including Joint Multimodal Aspect-Sentiment Analysis (JMASA), Multimodal Aspect Term Extraction (MATE), and Multimodal Aspect-oriented Sentiment Classification (MASC). 
We model these downstream tasks based on the same BART-based generative framework in vision-language pre-training, so that the downstream task can benefit more from pre-training during the fine-tuning stage. 
Following \citet{yan2021unified}, we formulate the outputs of the three subtasks as follows:
%which share the same architecture with pretraining framework 
\begin{itemize}
\setlength{\itemsep}{0pt}
\item
JMASA: $Y=[a_1^s,a_1^e,s_1,...,a_i^s,a_i^e,s_i,...]$,
\item
MATE: $Y=[a_1^s,a_1^e,...,a_i^s,a_i^e,...]$,
\item
MASC: $Y=[\underline{a_1^s},\underline{a_1^e},s_1,...,\underline{a_i^s},\underline{a_i^e},s_i,...]$,
\end{itemize}
where $a^s_i$, $a^e_i$, and $s_i$ inform the start index, end index, and sentiment of an aspect term in the text. The underlined tokens are given during inference.

Similar to the AOE task in Section~\ref{text_pretrain}, we formulate all the subtasks as index generation tasks, and use Eqn.~(\ref{eqn2}) to Eqn.~(\ref{eqn4}) to generate the token distribution.
The difference is that the special token set is modified as  $C=[\langle\textit{POS}\rangle,\langle\textit{NEU}\rangle,\langle\textit{NEG}\rangle,\langle\textit{EOS}\rangle]$ by adding the sentiment categories.
Figure~\ref{Fig.downstream} shows an example for JMASA.
Since the aspect-sentiment pairs are (\textit{Sergio Ramos}, \textit{Positive}) and (\textit{UCL}, \textit{Neutral}), its target sequence is $[1,2,\langle\textit{POS}\rangle,9,9,\langle\textit{NEU}\rangle,\langle\textit{eos}\rangle]$.
%The formulation of the subtasks is similar to the AOE task in pre-training. The difference is that the special token set used for downstream tasks is $C=[\textit{POS},\textit{NEU},\textit{NEG},\textit{<eos>}]$ which adds the types of sentiment. Specifically, Figure~\ref{Fig.downstream} displays an example of JMASA. The aspects in the text are \textit{Sergio Ramos} and \textit{UCL}, and the corresponding sentiments are \textit{positive} and \textit{neutral}. So the target sequence is $[1,2,\textit{POS},9,9,\textit{NEU},\textit{<eos>}]$.
%AOE中使用的fuctional token list,这里变为$C=[<POS>,<NEU>,<NEG>]$表示sentiment类别.

% \end{CJK}
% \begin{CJK}{UTF8}{gbsn}
\section{Experiment}
% In this section, we conduct different sets of experiments with
% the aim of answering the following research questions:
% \begin{itemize}
% \item
% RQ1: Can our pretraining framework outperform the state-of-the-art approach and other strong baselines over JMASA task?
% \item
% RQ2: Can our pretraining framework be helpful for two subtasks MATE and MASC?
% \item
% RQ3: What's the effectiveness of each pretraining task in our framework?
% \item
% RQ4: What's  the performance when we use different scale downstream dataset under our pretraining framework?
% \end{itemize}

\begin{table}[!tp]
\centering
\scriptsize
\setlength{\tabcolsep}{2.3mm}
\setlength{\belowcaptionskip}{-1em}
\begin{tabular}{l|ccc|ccc}
\toprule
&\multicolumn{3}{c|}{TWITTER-2015}&\multicolumn{3}{c}{TWITTER-2017} \\
\cmidrule(r){2-4} \cmidrule(r){5-7}
& \textbf{Train} & \textbf{Dev} & \textbf{Test}  & \textbf{Train} & \textbf{Dev} & \textbf{Test} \\ \midrule\midrule
Positive     &928	    &303	&317	&1508	&515	&493 \\ 
Neutral      &1883	&670	&607	&1638	&517	&573 \\
Negative     &368	    &149	&113	&416	&144	&168 \\  \midrule
Total Aspects      &3179	&1122	&1037	&3562	&1176	&1234 \\ 
\midrule\midrule
\text{\#Sentence} &2101 &727 &674 &1746 &577 &587\\
\bottomrule
\end{tabular}
\caption{\footnotesize The basic statistics of  two TWITTER datasets.}
\label{tab:data_statistics}
\end{table}

\subsection{Settings}
% We employ BART-base~\cite{lewis2020bart}作为我们的整体框架结构,encoder和decoder均为固定的6层, 并用其参数作为初始化. 预训练和下游batch size分别为64和16,epoch分别为40和35,采用AdamW来优化,warmup为0.1.encoder和decoder的隐层均为768维.We implement all the models with PyTorch, and run experiments on a RTX3090 GPU.
\textbf{Downstream datsets.} We adopt two benchmark datasets annotated by \citet{yu2019adapting}, namely TWITTER-2015 and TWITTER-2017 to evaluate our model. The statistics of the two datasets are shown in Table~\ref{tab:data_statistics}.

\textbf{Implementation Details.} 
We employ BART-base~\cite{lewis2020bart} as our framework. Specifically, the encoder and decoder both have six layers and are initialized with BART-base parameters. 
We fix all the hyper-parameters after tuning them on the development set.
The pre-training tasks were trained for 40 epochs and the downstream tasks were fine-tuned for 35 epochs.  
The batch sizes are set to 64 and 16, respectively. 
The learning rate is set to 5e-5.
The hidden size of our model is set to 768, which is the same as BART. 
The tradeoff hyper-parameters $\lambda_1$, $\lambda_2$, $\lambda_3$, $\lambda_4$, and $\lambda_5$ are all set to 1.
Note that for the subtask MASC, different from~\citet{ju2021emnlp} evaluating on the correctly predicted aspects, we provide all the golden aspects to the decoder of our framework during the inference stage and evaluate on all the aspects. %optimize both aspects and sentiments during training and
We implement all the models with PyTorch, and run experiments on a RTX3090 GPU.

%For downstream tasks, we 
%We first perform pre-training and then finetune on downstream datasets. The model with the most outstanding performance on the validation set will be saved, and then we evaluate its performance on the test set, which is the final results of our approach.

\textbf{Evaluation Metrics.}
% Following previous studies, 我们使用我们的预训练模型在下游进行finetuning后,选取在dev上最好的模型在test上测试作为我们的结果.
% 对于MABSA和MATE,我们采用micro F1 measure (F1), Precision(P) and Recall(R);对于MASC任务,我们使用了micro F1 measure (F1)和Accuracy(Acc).
We evaluate our model over three subtasks of MABSA and adopt Micro-F1 score (\text{F1}), Precision (\text{P}) and Recall (\text{R}) as the evaluation metrics to measure the performance. For MASC, to fairly compare with other approaches, we also use Accuracy (\text{Acc}).

% \begin{figure}[htbp]
% \subfigure[]
% {\begin{minipage}[b]{0.48\textwidth}
% \centering
%  \includegraphics[scale=0.20]{./fig/15_sample_add.pdf}
%   \end{minipage}
% }
% \subfigure[]
% {\begin{minipage}[b]{0.48\textwidth}
% \centering
%  \includegraphics[scale=0.2]{./fig/17_sample_add.pdf}
% %\caption{fig2}
% \end{minipage}%
% }
% \end{figure}

\subsection{Compared Systems}
%写法要和苏大区别开来!!!!
In this section, we introduce four types of compared systems for different tasks.

\textbf{Approaches for Multimodal Aspect Term Extraction (MATE).} 
1) \textit{RAN}~\cite{wu2020multimodal}, which aligns text with object regions by a co-attention network. 
2) \textit{UMT}~\cite{yu2020improving}, which uses Cross-Modal Transformer to fuse text and image representations for Multimodal Named Entity Recognition (MNER).
3) \textit{OSCGA}~\cite{wu:2020mm}, another MNER approach using visual objects as image representations. 
4) \textit{RpBERT}~\cite{sun2021rpbert}, which uses a multitask training model for MNER and image-text relation detection.

% \noindent
\textbf{Approaches for Multimodal Aspect Sentiment Classification (MASC).} 
1) \textit{TomBERT}~\cite{yu2019adapting}, which tackles the MASC task by employing BERT to capture intra-modality dynamics. 
2) \textit{CapTrBERT}~\cite{khan2021exploiting}, which translates the image to a caption as an auxiliary sentence for sentiment classification.

% \noindent
\textbf{Text-based approaches for Joint Aspect-Sentiment Analysis (JASA).} 
1) \textit{SPAN}~\cite{hu2019open}, which formulates the JASA task as a span prediction problem. 
2) \textit{D-GCN}~\cite{coling/ChenTS20}, which proposes a directional graph convolutional network to capture the correlation between words. 3) \textit{BART}~\cite{yan2021unified}, which adapts the JASA task to BART by formulating it as an index generation problem.
% and the results will be shown to compare with the above text-based methods. Note that our \textbf{T-only} does not use the task-specific pre-training.?????

\begin{table}[!tp]
\centering
\scriptsize
\begin{tabular}{l|ccc|ccc}
\toprule
  & \multicolumn{3}{c|}{TWITTER-2015}                                      & \multicolumn{3}{c}{TWITTER-2017}                                      \\ %\multicolumn{1}{l|}{\multirow{3}{*}{Methods}}
  \cmidrule(r){2-4}  \cmidrule(r){5-7}
 \multicolumn{1}{l}{}          
& \multicolumn{1}{|c}{\textbf{P}} & \multicolumn{1}{c}{\textbf{R}} & \multicolumn{1}{c}{\textbf{F1}} & \multicolumn{1}{|c}{\textbf{P}} & \multicolumn{1}{c}{\textbf{R}} & \multicolumn{1}{c}{\textbf{F1}} \\
\midrule\midrule  
\multicolumn{7}{l}{Text-based methods}\\
\midrule
 SPAN$^{*}$  & 53.7  & 53.9  & 53.8  & 59.6  & 61.7  & 60.6 \\
 D-GCN$^{*}$ & 58.3  & 58.8  & 59.4  & 64.2  & 64.1  & 64.1 \\
 BART  & 62.9  & 65.0  & 63.9  & 65.2 & 65.6 & 65.4  \\
\midrule\midrule
\multicolumn{7}{l}{Multimodal methods}\\
\midrule
UMT+TomBERT$^{*}$  & 58.4  & 61.3   & 59.8 & 62.3  & 62.4 & 62.4 \\
OSCGA+TomBERT$^{*}$  & 61.7 & 63.4 & 62.5 & 63.4 & 64.0  & 63.7\\
OSCGA-collapse$^{*}$  & 63.1 & 63.7 & 63.2 & 63.5   & 63.5 & 63.5 \\
RpBERT-collapse$^{*}$ & 49.3 &46.9 &48.0 &57.0 &55.4 &56.2\\
JML$^{*}$ & 65.0  & 63.2  & 64.1 & 66.5 & 65.5 & 66.0\\
VLP-MABSA  & \textbf{65.1}  & \textbf{68.3} & \textbf{66.6} & \textbf{66.9}  & \textbf{69.2}  & \textbf{68.0} \\
\bottomrule                            
\end{tabular}
\caption{\footnotesize Results of different approaches for JMASA. $^{*}$ denotes the results are from~\citet{ju2021emnlp}.}
\label{tab:JMASA}
\end{table}

% \noindent
\textbf{Multimodal approaches for Joint Multimodal Aspect-Sentiment Analysis (JMASA).} 
%By combining methods for subtasks mentioned above, some pipeline approaches are used to solve JMASA
1) \textit{UMT+TomBERT} and \textit{OSCGA+TomBERT}, which are simple pipeline approaches by combining methods for subtasks mentioned above.
2) \textit{UMT-collapsed}~\cite{yu2020improving}, \textit{OSCGA-collapsed}~\cite{wu:2020mm} and \textit{RpBERT-collapsed}~\cite{sun2021rpbert}, which model the JMASA task with collapsed labels such as \textit{B-POS} and \textit{I-POS}. 
3) \textit{JML}~\cite{ju2021emnlp}, which is a multi-task learning approach proposed recently with the auxiliary cross-modal relation detection task.

\subsection{Main Results}
In this section, we analyze the results of different approaches on three subtasks of MABSA.

\textbf{Results of JMASA.} 
% 从表4展示的结果可以发现text-based approaches 比使用多模态方法效果要差不少,这说明图像信息可以为这个任务提供很大的帮助,besides \textbf{Ours-text}显著超越了其他text方法,例如在TWITTER-2015上比D-GCN$F_1$高了4.5,这证明了我们framework的优越性;for multimodal approaches,\textbf{JML}显著超越了之前的所有方法,说明使用的relation detection辅助任务和joint framework is beneficial.我们的方法比SOTA的JML在TWITTER-2015上高了3.3,在TWITTER-2017上高了1.7这得益于我们所使用的aspect and opinion aware的预训练任务,挖掘了image和text中的主客观信息;
% most text-based methods perform much worse than multimodal methods. This suggests that visual modality can provide extra clues for the task. Besides, we can find that
Table~\ref{tab:JMASA} shows the results of different methods for JMASA. As we can see from the table, \textit{BART} achieves the best performance among text-based methods, and it even outperforms some multimodal methods, which proves the superiority of our base framework. 
For multimodal methods, \textit{JML} achieves better performance than previous methods mainly due to its auxiliary task about relation detection between image and text. Among all the methods, \textit{VLP-MABSA} which is the whole model with all the pre-training tasks consistently performs the best across two datasets. %For instance, on TWITTER-2015 our method exceeds JML by 1.5\%,5.7\%,3.3\% with respect to \textit{Precision}, \textit{Recall} and \textit{F1 score}, respectively.
Specifically, it significantly outperforms the second best system \textit{JML} with 2.5 and 2.0 absolute percentage points with respect to \textit{F1} on TWITTER-2015 and TWITTER-2017, respectively.
This mainly benefits from our task-specific pre-training tasks, which identify aspects and opinions as well as their alignments across the two modalities.

\begin{table}[!t]
\centering
\scriptsize
\begin{tabular}{l|ccc|ccc}
\toprule
\multirow{2}{*}{Methods}  & \multicolumn{3}{c|}{TWITTER-2015} & \multicolumn{3}{c}{TWITTER-2017}                                      \\\cmidrule(r){2-4}  \cmidrule(r){5-7}
& \multicolumn{1}{|c}{\textbf{P}} & \multicolumn{1}{c}{\textbf{R}} & \multicolumn{1}{c}{\textbf{F1}} & \multicolumn{1}{|c}{\textbf{P}} & \multicolumn{1}{c}{\textbf{R}} & \multicolumn{1}{c}{\textbf{F1}} \\
\midrule\midrule                            
RAN$^{*}$  & 80.5  & 81.5    & 81.0 & 90.7    & 90.0 & 90.3  \\
UMT$^{*}$ & 77.8  & 81.7    & 79.7  & 86.7    & 86.8 & 86.7  \\
OSCGA$^{*}$  & 81.7 & 82.1     & 81.9 & 90.2   & 90.7 & 90.4  \\
JML-MATE$^{*}$ & 83.6 & 81.2 & 82.4 & \textbf{92.0} & 90.7 & 91.4 \\
VLP-MABSA & \textbf{83.6} & \textbf{87.9} & \textbf{85.7} & 90.8 & \textbf{92.6} & \textbf{91.7}\\
\bottomrule                            
\end{tabular}
\caption{\footnotesize Results of different approaches for MATE. $^{*}$ denotes the results are from~\citet{ju2021emnlp}.}
\label{tab:MATE}
\end{table}

\begin{table}[!t]
\centering
\scriptsize
\setlength{\belowcaptionskip}{-1em}
\begin{tabular}{l|ccc|ccc}
\toprule
\multirow{2}{*}{Methods}  & \multicolumn{3}{c|}{TWITTER-2015} & \multicolumn{3}{c}{TWITTER-2017}\\ \cmidrule(r){2-4}  \cmidrule(r){5-7}
& \textbf{Acc} & &\textbf{F1} & \textbf{Acc} & &\textbf{F1}\\
\midrule\midrule                            
TomBERT  & 77.2 & &71.8 & 70.5 & &68.0  \\
CapTrBERT & 78.0 & &73.2 & 72.3 & &70.2   \\
JML-MASC  & \textbf{78.7} & &- & 72.7 & &- \\
VLP-MABSA & 78.6 & &\textbf{73.8} &\textbf{73.8} & &\textbf{71.8} \\                            
\bottomrule                            
\end{tabular}
\caption{\footnotesize Results of different approaches for MASC. Note that JML-MASC only evaluates on the aspects correctly predicted by JML-MATE while the other methods evaluate on all the golden aspects.}
\label{tab:MASC}
\end{table}

% \noindent
\textbf{Results of MATE and MASC.}
% 表5展示了不同方法在MATE子任务上的performance. JML的效果超过了其他的多模态方法, 说明其设计的辅助任务能够有一定的helpful.我们的方法也极大超越了JML,特别是在在TWITTER-15上有接近4个点的提升, 这反映出基于我们的框架结合task-specific的预训练任务通过挖掘image和text中的客观信息可以很好地帮助模型进行aspect term的抽取.
% 表6展示了不同方法在MASC子任务上的performance.我们的方法也取得了不错的效果,其中在TWITTER17上超越了JML1个点,这得益于我们设计的预训练任务挖掘image and text中的subjective information的能力对于image-text pair级别sentiment的理解.
Table~\ref{tab:MATE} and Table~\ref{tab:MASC} show the results of MATE and MASC, respectively. 
Similar to the trend on the JMASA subtask, we can clearly observe that our proposed approach \textit{VLP-MABSA} generally achieves the best performance across the two datasets, except on the accuracy metric of TWITTER-2015.
These observations further demonstrate the general effectiveness of our proposed pre-training approach.
%all the previous baselines perform worse than \textit{Ours-MATE}. This is mainly because the ability to capture objective information from image and text of these methods is not strong enough, which our task-specific pre-training tasks can enhance, especially the AOE and the AOG task.????

%\textbf{Results of MASC.}
% shows the results of MASC. Our method achieves outstanding performance compared with previous works. This is probably because our task-specific pre-training tasks successfully capture the subjective information from two modalities.????

\subsection{In-depth Analysis of Pre-training Tasks}

To explore the impact of each pre-training task, we perform a thorough ablation study over the full supervision setting which uses full training dataset and the weak supervision setting which only randomly chooses 200 training samples for fine-tuning. 

\begin{table}[!tp]
\centering
\scriptsize
%\footnotesizehttps://www.overleaf.com/project/616798e5cadd6b653c9329ea
\setlength{\belowcaptionskip}{-1.2em}
\setlength{\tabcolsep}{1mm}
{\begin{tabular}{clccc|ccc}
\toprule
&&\multicolumn{3}{|c|}{TWITTER-2015}                                  & \multicolumn{3}{c}{TWITTER-2017}  \\
\cmidrule{3-5}\cmidrule{6-8}
                     && \multicolumn{1}{|c}{JMASA} & \multicolumn{1}{c}{MATE} & \multicolumn{1}{c|}{MASC} & \multicolumn{1}{c}{JMASA} & \multicolumn{1}{c}{MATE} & \multicolumn{1}{c}{MASC} \\
%& Pretraining Task(s) & P       & R      & $F_1$      & P      & R      & $F_1$      & ACC               & P       & R      & $F_1$     & P      & R      & $F_1$      & ACC        \\
\midrule\midrule
% \multirow{6}{*}{\rotatebox{90}{supervised}} &
% \midrule
\multirow{6}{*}{\rotatebox{90}{Full supervision}}&\multicolumn{1}{|l|}{w/o pre-training}          &65.31        &84.80        &76.81                     &66.10                &90.67       &72.78                             \\
&\multicolumn{1}{|l|}{+\textbf{T}$_{\text{MLM}}$}             &65.44               &84.91        &77.08              &66.27           &91.00         &72.82     
\\
&\multicolumn{1}{|l|}{+\textbf{T}$_{\text{AOE}}$}       &65.92        &85.43        &77.48         &67.12      &91.75        &72.89                               \\
&\multicolumn{1}{|l|}{+\textbf{V}$_{\text{MRM}}$}           &65.94       &85.49        &77.53                   &67.15        &91.72        &73.13                              \\
&\multicolumn{1}{|l|}{+\textbf{V}$_{\text{AOG}}$}      &66.38      &\textbf{85.73}        & 77.82          &67.66 &\textbf{91.77}        & 73.32                                \\
&\multicolumn{1}{|l|}{ + \textbf{MM}$_{\text{MSP}}$}           &\textbf{66.64}      &85.66        &\textbf{78.59}              & \textbf{68.05}          &91.73         &\textbf{73.82}                         \\

\midrule\midrule

% \multirow{6}{*}{\rotatebox{90}{weakly-supervised}} &
% \midrule
\multirow{6}{*}{\rotatebox{90}{Weak supervision}}&\multicolumn{1}{|l|}{w/o pre-training}          &39.79         &69.33              &57.40                 &49.12                &80.48       &61.04                         \\
&\multicolumn{1}{|l|}{+\textbf{T}$_{\text{MLM}}$}           &40.42       &69.69        &58.00         &49.69      &81.26        &61.15       
\\
&\multicolumn{1}{|l|}{+\textbf{T}$_{\text{AOE}}$}      &46.15        &79.13        &58.32               &52.00        &84.60        &61.46                        \\
&\multicolumn{1}{|l|}{+\textbf{V}$_{\text{MRM}}$}         &46.64        &79.49        &58.68               &52.18         &84.47        &61.78                      \\
&\multicolumn{1}{|l|}{+\textbf{V}$_{\text{AOG}}$}       &47.79        &\textbf{80.94}              &59.32                &53.16       &\textbf{85.04}        &62.51                      \\
&\multicolumn{1}{|l|}{+ \textbf{MM}$_{\text{MSP}}$}      &\textbf{51.71}          &80.69        &\textbf{62.58}                            &\textbf{55.38}          &84.88        &\textbf{64.42}   \\

\bottomrule
\end{tabular}}

\caption{\footnotesize The results of pre-training tasks on two benchmarks. We evaluate over three tasks JMASA, MATE, and MASC in terms of \textit{F1}, \textit{F1} and \textit{Acc}, respectively. \textit{T}, \textit{V}, and \textit{MM} denote the Textual, Visual, and Multimodal pre-training, respectively. Each row adds an extra pre-training task to the row above it.}
\label{tab:pretraining}
\end{table}

\textbf{Impact of Each Pre-training Task.}
As we can see from Table~\ref{tab:pretraining}, the performance generally improves with respect to most metrics when adding more pre-training tasks. 

To better analyze the effect of each pre-training task, we take the weak supervision experiments on TWITTER-2015 as an example. 
When only using MLM to pre-train our model, the performance only gets slight improvements. 
After adding the AOE task, the result of MATE gets a huge improvement of 9.44\% on \textit{F1}. 
This shows that the AOE task greatly enhances our model's ability to recognize the aspect terms. 
When adding the MRM task, the performance gets slight improvements again. This reflects that general pre-training tasks (e.g., MLM and MRM) are not adequate for our model to tackle downstream tasks which need the model to understand the subjective and objective information from image and text. 
When adding the AOG task, the performance over three subtasks gets a moderate improvement, which proves the effectiveness of the AOG task.
Finally, adding the MSP task significantly boosts the performance, especially on the MASC task. This 
shows that the MSP task can enhance our model's understanding of sentiment across language and image modalities. 
By combining all the pre-training tasks, our full model generally achieves the best results over most of the subtasks whether in both full supervision and weak supervision settings.

\begin{figure}
  \setlength{\belowcaptionskip}{-0.3cm}
  \setlength{\abovecaptionskip}{0.1cm}
  \scriptsize
\begin{tabular}{p{3.5cm}p{3.5cm}}
\begin{minipage}{0.1\textwidth}
% \centering
    \hbox{\hspace{-2em} \includegraphics[width=42mm, height=38.0mm]{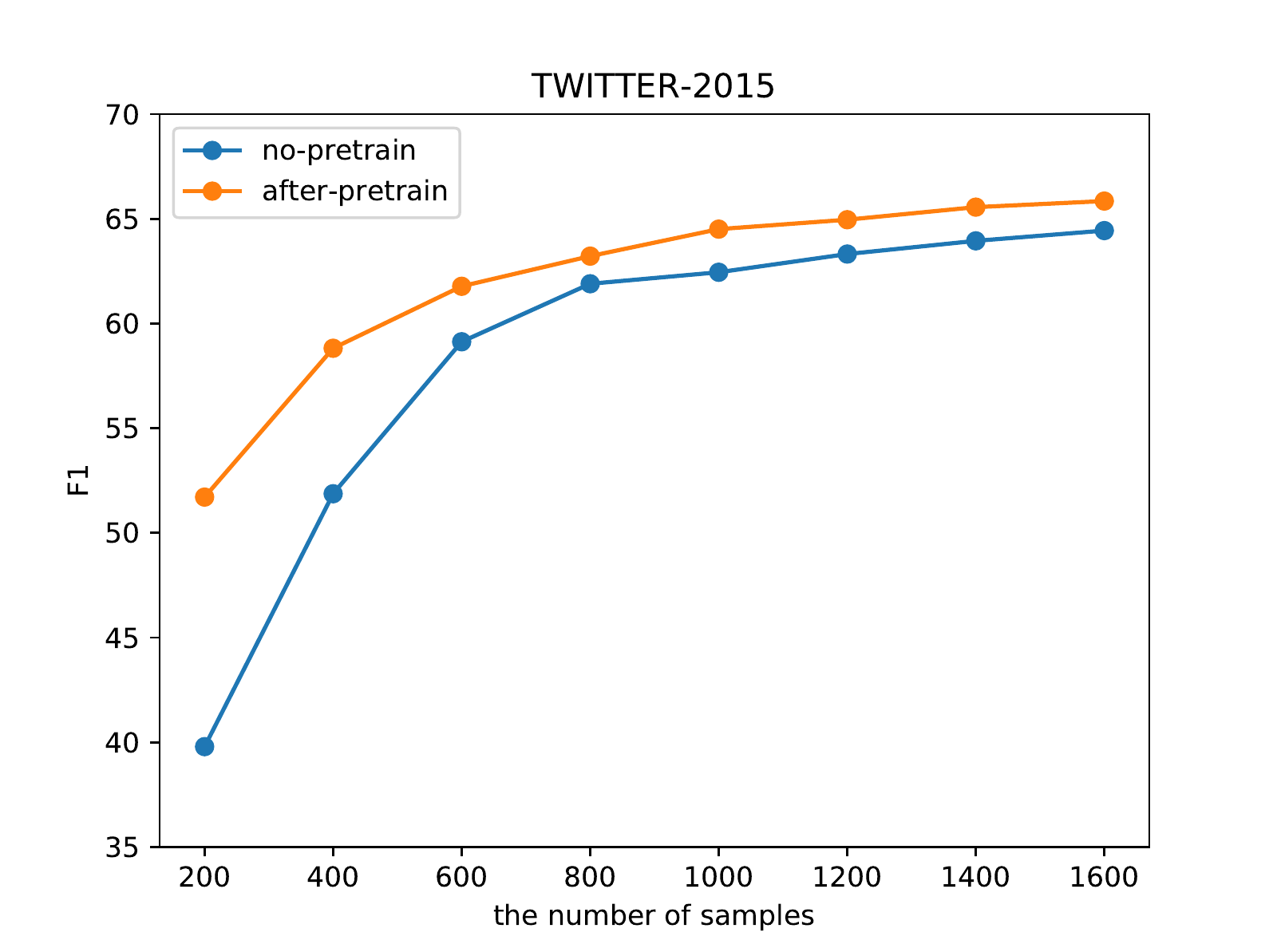}}
  \end{minipage}\vspace{0.3em}
&
\begin{minipage}{0.1\textwidth}
% \centering
    \hbox{\hspace{-1.8em} \includegraphics[width=42mm, height=38.0mm]{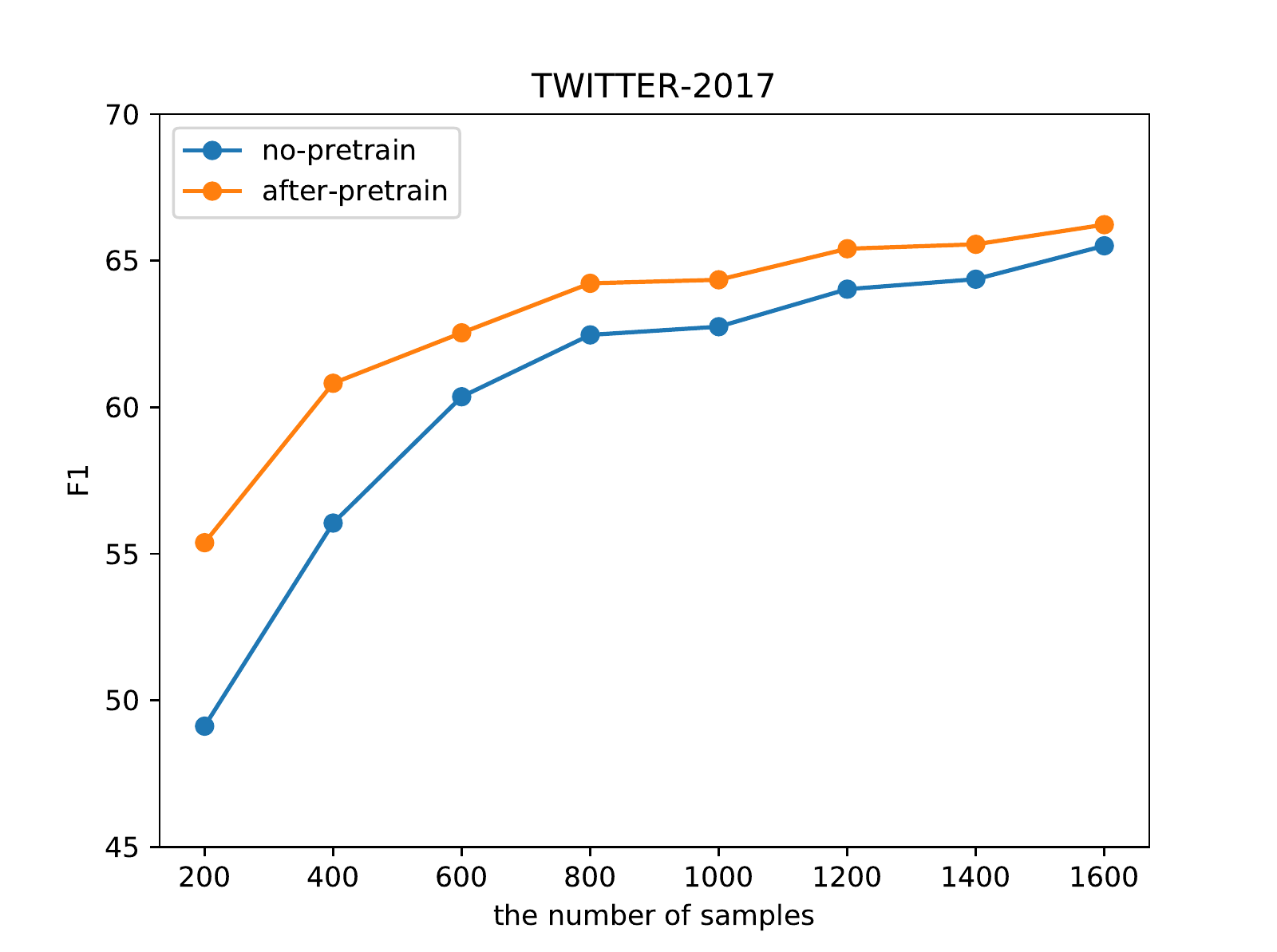}}
  \end{minipage}\vspace{0.3em}\\
\end{tabular}
\caption{\footnotesize The effectiveness of pre-training when using different number of training samples for the downstream task. Y-axis refers to the \textbf{F1} score (\%) of the JMASA task.}
\label{fig:sample_num}
\end{figure}

\textbf{Impact of pre-training when using different number of downstream training samples.}
% 图4展示了利用不同数量样本进行下游训练时，使用和不使用预训练的效果。从图中可以看出,相比不使用预训练，使用预训练均可以带来一定的提升.当样本数量较少时，使用预训练可以获得非常大的提升,说明当下游的知识较少时，预训练可以提供比较大的帮助;随着样本数量的增加，预训练的效果开始变得相对较弱,但也可以获得一定的帮助。
To better understand the impact of pre-training, we compare the results with and without pre-training when adopting different number of samples for downstream training. 
We use the JMASA task as the example to observe the impact. 
As shown in Fig.~\ref{fig:sample_num}, when the sample size is small, pre-training can bring a huge improvement. 
In contrast, when the sample size becomes larger, pre-training brings relatively small improvements. This further illustrates the robustness and the effectiveness of our pre-training approach, especially in low-resource scenarios.

\subsection{Case Study}

\begin{table*}[!t]
  \setlength{\belowcaptionskip}{-0.3cm}
  \setlength{\abovecaptionskip}{0.1cm}
  \scriptsize
\begin{tabular}{p{0.6cm}p{3.4cm}p{3.3cm}p{3.3cm}p{3.2cm}}
\toprule
Image &
\begin{minipage}{0.1\textwidth}
    \hbox{\hspace{-0.1em} \includegraphics[width=34mm, height=25.0mm]{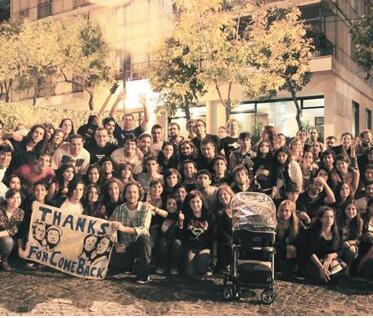}}
  \end{minipage}\vspace{0.3em}
&
\begin{minipage}{0.1\textwidth}
    \hbox{\hspace{-0.1em} \includegraphics[width=32mm, height=25.0mm]{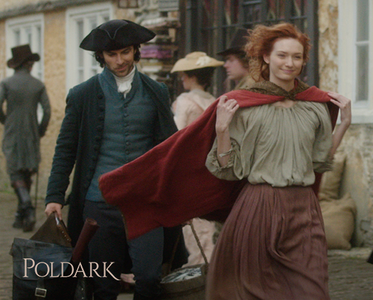}}
  \end{minipage}\vspace{0.3em}
&
\begin{minipage}{0.1\textwidth}
    \hbox{\hspace{-0.1em} \includegraphics[width=32mm, height=25.0mm]{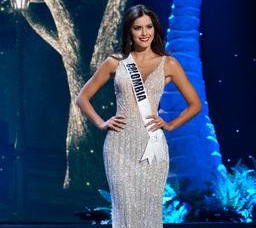}}
  \end{minipage}\vspace{0.3em}
&
\begin{minipage}{0.1\textwidth}
    \hbox{\hspace{-0.1em} \includegraphics[width=32mm, height=25.0mm]{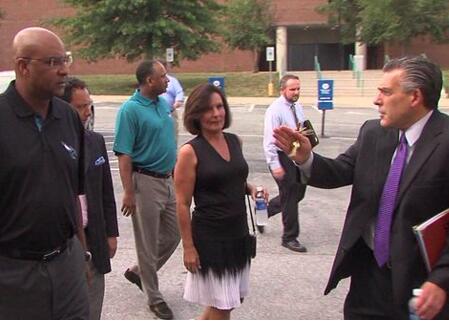}}
  \end{minipage}\vspace{0.3em}
\\
%\toprule
\multirow{4}{*}{Text} &
(a) RT @ PearlJam : Eddie and the Faithfull Pearl Jam fans in Buenos Aires . Photo by @ epozzoni \# PJSA2013
&
(b) RT @ BBCOne : Dear Madonna , THIS is how you wear a cape . \# Poldark \# Demelza
&
(c)  RT @ TrumpDoral : Congratulations to the the new \# MissUniverse , Miss Colombia , Paulina Vega ! 
&
(d) RT @ myfox8 : Charlotte @ hornets visit \# Greensboro for D - League meeting
\\
\midrule
\multirow{3}{*}{GT} &
\multirow{1}{*}{(Eddie, POS)} & (Madonna, POS) & (Miss Colombia, POS) & (Charlotte, NEU)\\
& (Pearl Jam, POS)&(Poldark, NEU) & (Paulina Vega, POS) & (Greensboro, NEU)\\
& (Buenos Aires, NEU) & (Demelza, NEU) & & (D – League, NEU)\\
\cmidrule{2-5}
\multirow{3}{*}{BART} & 
\multirow{1}{*}{(Eddie, NEU)~$\times$}  & (Madonna, POS)~$\checkmark$ & (Colombia, POS)~$\times$ & (Charlotte, NEU)~$\checkmark$\\
& (the Faithfull Pearl Jam, NEU)~$\times$ & ~~~~~~~~~~~~-~$\times$ & (Paulina Vega, POS)~$\checkmark$ & (Greensboro, NEU)~$\checkmark$\\
& (Buenos Aires, NEU)~$\checkmark$& ~~~~~~~~~~~~-~$\times$ & & ~~~~~~~~~~~~-~$\times$ \\\cmidrule{2-5}
\multirow{3}{*}{MM} & 
\multirow{1}{*}{(Eddie, NEU)~$\times$}  & (Madonna, NEU)~$\times$ & (Colombia, NEU)~$\times$ & (Charlotte, NEU)~$\checkmark$\\
& (the Faithfull Pearl Jam, NEU)~$\times$ & ~~~~~~~~~~~~-~$\times$ & (Paulina Vega, POS)~$\checkmark$ & (Greensboro, NEU)~$\checkmark$\\
& (Buenos Aires, NEU)~$\checkmark$ & (Demelza, NEU)~$\checkmark$ & & ~~~~~~~~~~~~-~$\times$\\\cmidrule{2-5} 
\multirow{3}{*}{VLP} & \multirow{1}{*}{(Eddie, POS)~$\checkmark$}  & (Madonna, POS)~$\checkmark$ & (Miss Colombia, POS)~$\checkmark$ & (Charlotte, NEU)~$\checkmark$\\
& (Pearl Jam, POS)~$\checkmark$ & (Poldark, NEU)~$\checkmark$ & (Paulina Vega, POS)~$\checkmark$ & (Greensboro, NEU)~$\checkmark$\\
& (Buenos Aires, NEU)~$\checkmark$& (Demelza, NEU)~$\checkmark$ & & (D – League, NEU)~$\checkmark$\\
\toprule
\end{tabular}
\caption{Predictions of different methods on four test samples. NEU, POS, and NEG denote Neutral, Positive, and Negative sentiments, respectively.}
%More case studies are presented in the Appendix.}
\label{tab:case_study}
\end{table*}
To further demonstrate the effectiveness of our approach, we present four test examples with predictions from different methods. 
The compared methods are \textit{BART}, our framework using multimodal inputs without pre-training (denoted by \textit{MM}), and our framework using multimodal inputs with full pre-training (denoted by \textit{VLP}), respectively. 
As shown in Table~\ref{tab:case_study}, for example (a), both \textit{BART} and \textit{MM} extracted the wrong aspect term (i.e., \textit{the Faithfull Pearl Jam}) and gave the incorrect sentiment prediction towards \textit{Eddie}. 
For example (b), \textit{BART} only extracted one aspect term \textit{Madonna} while \textit{MM} identified an additional aspect term \textit{Demelza}. However, the sentiment towards \textit{Madonna} was wrongly predicted by \textit{MM}. 
For example (c), \textit{BART} only recognized part of the aspect term \textit{Colombia} and \textit{MM} wrongly predicted the sentiment towards \textit{Miss Colombia} as \textit{Neutral}. 
For example (d), both \textit{BART} and \textit{MM} failed to recognize the aspect term \textit{D-League}.
Among all the cases, our \textit{VLP} model with full pre-training correctly extracted all the aspect terms and classified the sentiment , which shows the advantage of our generative framework and task-specific pre-training tasks.
% \end{CJK}
\section{Conclusion}
In this paper, we proposed a task-specific Vision-Language Pre-training framework for Multimodal Aspect-Based Sentiment Analysis (VLP-MABSA). We further designed three kinds of pre-training tasks from the language, vision, and multi-modal modalities, respectively. 
%Via these task-specific pre-training tasks, we could extract fine-grained objective and subjective information as well as their relations from texts and images. 
Experimental results show that our proposed approach generally outperforms the state-of-the-art methods for three subtasks of MABSA.
Our work is a first step towards a unified Vision-Language Pre-training framework for MABSA.
In the future, we plan to apply our pre-training approach on a larger dataset and consider the relation between image and text in our pre-training framework.
We hope this work can potentially bring new insights and perspectives to the research of MABSA.
%In the future, we plan to apply our pre-training approach on a larger dataset. Besides, we also plan to consider the relation between image and text in our pre-training framework.

\section*{Acknowledgments}
The authors would like to thank the anonymous reviewers for their insightful comments. 
%Jianfei Yu and Rui Xia are the corresponding authors.
This work was supported by the Natural Science Foundation of China (62076133 and 62006117), and the Natural Science Foundation of Jiangsu Province for Young Scholars (BK20200463) and Distinguished Young Scholars (BK20200018).

%\section*{Acknowledgements}

%T \emph{International Joint Conference on Artificial Intelligence} and the \emph{Conference on Computer Vision and Pattern Recognition}.

\bibliographystyle{acl_natbib}
\bibliography{reference}

\end{document}